\pdfoutput=1

\documentclass[11pt]{article}

\usepackage[preprint]{acl}

\usepackage{times}
\usepackage{latexsym}

\usepackage[T1]{fontenc}

\usepackage[utf8]{inputenc}

\usepackage{microtype}

\usepackage{inconsolata}

\usepackage{algorithm}
\usepackage{algpseudocode}
\usepackage{amsmath}
\usepackage{amsfonts}

\usepackage{booktabs} 
\usepackage{subcaption}
\usepackage{graphicx}
\usepackage{hyperref}

\title{OptBA: Optimizing Hyperparameters with the Bees Algorithm for Improved Medical Text Classification}

\author {
    Mai A. Shaaban\textsuperscript{\rm 1,\rm 2} \and
    Mariam Kashkash\textsuperscript{\rm 1} \and
    Maryam Alghfeli\textsuperscript{\rm 1} \and
    Adham Ibrahim\textsuperscript{\rm 1} \\
    \textsuperscript{\rm 1}Mohamed bin Zayed University of Artificial Intelligence, U.A.E\\
    \textsuperscript{\rm 2}Faculty of Science, Alexandria University, Egypt\\
    \text{\{mai.kassem, mariam.kashkash, maryam.alghfeli, adham.ibrahim\}@mbzuai.ac.ae}
}

\begin{document}
\maketitle
\begin{abstract}
One of the main challenges in the field of deep learning is obtaining the optimal model hyperparameters. The search for optimal hyperparameters usually hinders the progress of solutions to real-world problems such as healthcare. Previous solutions have been proposed, but they can still get stuck in local optima. To overcome this hurdle, we propose OptBA to automatically fine-tune the hyperparameters of deep learning models by leveraging the Bees Algorithm, which is a recent promising swarm intelligence algorithm. In this paper, the optimization problem of OptBA is to maximize the accuracy in classifying ailments using medical text, where initial hyperparameters are iteratively adjusted by specific criteria. Experimental results demonstrate a noteworthy enhancement in accuracy with approximately 1.4\%. This outcome highlights the effectiveness of the proposed mechanism in addressing the critical issue of hyperparameter optimization and its potential impact on advancing solutions for healthcare. The code is available publicly at \url{https://github.com/Mai-CS/OptBA}.
\end{abstract}

\section{Introduction}
In the recent past, the expansion of the COVID-19 pandemic has reshaped the world radically. Hospitals and medical centres have become fertile ground for the spread of this virus. Social distancing plays a pivotal role in eliminating the spread of this virus \cite{Lotfi2020}. Moreover, the doctors' productivity may decrease due to the intense effort required to balance between in-patients and out-patients \cite{Wu2019}. Consequently, a wide variety of deep learning paradigms are applied to speed up the diagnosis process  \cite{bakator2018deep}. The aim of this work is to use natural language processing (NLP) models along with swarm intelligence algorithms such as the Bees Algorithm \cite{pham2006bees} for more accurate predictions.

The used English dataset \cite{Mooney} contains more than 6000 records of variant symptoms described by patients as free text along with the type of the ailment. The first step in the proposed work is to perform text preprocessing techniques such as lemmatization, stop word removal, and generating word embeddings. Then, a Long Short-Term Memory (LSTM) \cite{yu2019review} deep neural network is suggested to take word embeddings as inputs to predict the output (i.e., the ailment). However, LSTM as a deep learning model suffers from the risk of getting stuck in local optima. This is because the values of weights are initialized randomly. Not only the weights but also the hyperparameters \cite{Alsaleh2021}. In the proposed work, the Bees Algorithm (BA) \cite{pham2006bees} is adopted to enhance the process of hyperparameter tuning of LSTM. BA is a population-based algorithm that mimics the behaviour of bees in foraging in nature \cite{Kashkash2022}. To the best of our knowledge and based on an extensive literature review, this work is the first to integrate the Bees Algorithm with deep learning for medical text classification.

\section{Related Work}
LSTM models have been shown to be capable of achieving remarkable text classification performance compared to other deep learning models. \cite{sherstinsky2020fundamentals, ALHAMOUD2022}. A dynamic deep ensemble model was proposed to classify the text into spam or legitimate. The features were extracted by convolutional and pooling layers \cite{Shaaban2022}. This study \cite{sentiment} compared three CNN and five RNN structures across 13 datasets for sentiment classification. They found that larger training datasets consistently improved classification performance across all models, with varying effects observed between input levels in CNN and RNN architectures.

The swarm-based evolutionary algorithms (EAs) \cite{piotrowski2017swarm} operate by utilizing a population of solutions rather than relying on a single solution at each iteration. A novel population-based search technique known as the Bees Algorithm was introduced in \cite{pham2006bees}. The authors demonstrated that BA is capable of converging to either the maximum or minimum of the objective function, effectively avoiding being trapped at local optima and outperforming other competing methods in terms of speed and accuracy.

In the field of hyperparameter optimization, this paper \cite{yang2020hyperparameter} demonstrated that Bayesian Optimization HyperBand (BOHB) is the preferred option for obtaining the optimal hyperparameters when randomly selected subsets adequately represent the dataset. Alternatively, for a small hyperparameter configuration space, Bayesian Optimization (BO) models are advised, while Particle Swarm Optimization (PSO) typically proves most effective for larger configuration spaces. Moreover, the authors in this paper \cite{akiba2019optuna} introduced Optuna and demonstrated the superiority of Optuna's convergence speed, scalability, and ease of integration by leveraging optimization techniques to streamline the hyperparameter optimization process. However, it depends on pruning hyperparameters in a sequential manner, which can lead to a slower search procedure.

Other applications involve EA for automatic training of machine learning algorithms. For instance, this paper \cite{sym13081347} employed BA for training deep Recurrent Neural Networks for sentiment classification. However, the approach differs from our work in its primary objectives and applications. The authors use BA for training the model to obtain the optimal model parameters (i.e., optimal weights), whereas our paper focuses on utilizing BA for hyperparameter optimization.

\section{Methods}

\subsection{Data Preprocessing}
\begin{figure}[!htbp]
\centering
    \includegraphics [width=\columnwidth]{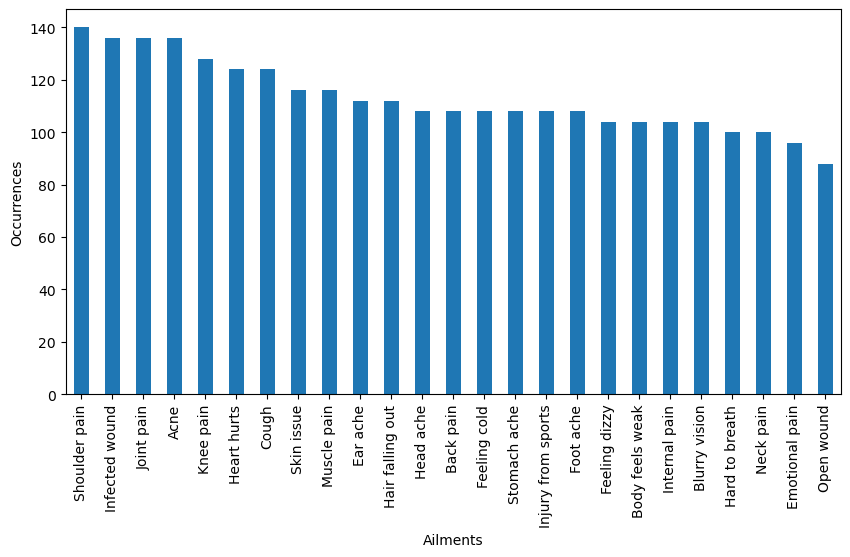}
    \caption{Class distribution of the ailments dataset.}
    \label{fig:class_distribution}
\end{figure}

It was observed that the data contained duplicated records and the number of records shrunk after dropping replicates to 706 data samples. Hence, we apply a text augmentation technique using the \textit{nlpaug} tool \cite{ma2019nlpaug} to enhance the performance and reduce the probability of overfitting. Consequently, the size of the data increased to 2829 text records along with 25 classes (ailments). Figure~\ref{fig:class_distribution} shows the balanced distribution of classes after applying the augmentation technique. Although there are several ways introduced for text augmentation, random swapping is chosen empirically based on the highest accuracy score.



We perform text preprocessing techniques, which involve tokenization, removal of stop words and lemmatization. First, each text is tokenized into words. Then, we remove stop words, which occur commonly across all texts. (e.g., "the", "is", "you"). Finally, we apply lemmatization for the sake of grouping different forms of the same word.

Converting textual data into digits is one of the main pillars of achieving natural language processing in various capacities \cite{mikolov2013efficient}. Therefore, the words must be expressed numerically to fit as inputs to deep learning models. The word embedding technique represents each word by a vector of numbers, indicating the semantic similarity between words. It creates a dense vector by transforming each word into a word vector that reflects its relative meaning within the document. The input is illustrated in a matrix $M \in \mathbb{R}^{n\times d}$, which denotes a collection of phrases. Each phrase $m \in M$ has a sequence of words: $w_1$,$w_2$,$w_3$, ...,$w_n$; and every word is represented in a word vector of length $d$ \cite{Shaaban2022}.








\subsection{The Bees Algorithm}
The Bees Algorithm (BA) is a swarm intelligence algorithm and a population-based algorithm that mimics the behaviour of honey bees in nature in order to forage \cite{pham2006bees}. In the beginning, scout bees are sent to discover the area. When those bees return, they perform a waggle dance that indicates the quality of the discovered batches. After that, recruiter bees are sent to the good batches to fetch good nectar, which enhances the quality and amount of produced honey.
BA is started by initializing the population of $n$ number of bees. After that, the main loop of BA is started by selecting $m$ good bees out of $n$ bees to implement the local search. The local search exploits the found solutions in order to reach the optimal one. The elite bees $e$ are selected out of $m$ bees, and they recruit $nep$ bees to help them find better solutions in their neighbourhood. While $nsp$ bees are recruited to search in the neighborhood of the remaining good bees ($m-e$). In general, $nep$ should be greater than $nsp$. The remaining bees in the population implement the global search to explore all available solutions. This loop is repeated until convergence.  

    

\subsubsection{Hyperparameter Tuning using the Bees Algorithm}
In this section, we introduce OptBA, a novel framework, in which BA is applied to find the optimal values of LSTM hyperparameters: the number of epochs required to train the LSTM model and the number of units in the LSTM layer. Thus, the structure of the bee in OptBA consists of the value of the number of epochs, the number of units and the accuracy value obtained from running LSTM. Considering that the remaining hyperparameters of LSTM are kept fixed during the experiment.

The algorithm starts by generating $n$ bees (solutions) randomly as the initial population, which represents $n$ different structures of LSTM. Each parameter is generated randomly using the Uniform distribution function.
After that, the evaluation function is implemented by training LSTM for each bee and the evaluation value is the obtained accuracy on the validation set. Next, these bees are ordered decently based on the resulting accuracy value. Then, the $m$ good bees are selected and the elite $e$ between them are distinguished. There are $nep$, and $nsp$ bees, which are recruited for each bee in the $m$ good bees and elite $e$, respectively, to enhance the found solution. This is performed by generating new values of the epoch and the unit parameters in the neighbourhood of original values by using the Uniform distribution function.

After implementing the local search, the global search is run for the remaining $(n-m)$ to discover new solutions that can be promised. The global search is implemented by replacing each remaining bee $(n-m)$  with a new one. Finally, the local search and the global search are repeated until the convergence or the maximum number of iterations of BA is reached. The detailed implementation of OptBA is in Algorithm~\ref{alg:OptBA}.   

\begin{algorithm}
\caption{The OptBA Algorithm}
\label{alg:OptBA}
\begin{algorithmic}[1]

	\State Input: $n$, $m$, $e$, $nep$, $nsp$, stopping criteria, word embeddings      
	\State Output: The optimal no. epochs and no. units
	\State Initialize a population of $n$ LSTM models.
	\State Train and evaluate each LSTM in the population.
	\While{stopping criteria are not satisfied}  
        \State Sort decently all models w.r.t their accuracy values.
        \State Select $m$ good models for local search.
        \State Select $e$ elite models for local search.
        \State Recruit $nep$ models for each of $e$ models.
        \State Recruit $nsp$ models for the remaining $m-e$ models.
        \State Send the remaining $n-m$ models for global search.
        \State Calculate the accuracy value for all models.
        \State Select the best model as optimal.
    \EndWhile
\end{algorithmic}
\end{algorithm}


\section{Results and Discussion}

\begin{table*}[!t]
  \centering
   \caption{Comparison of results of LSTM with default and optimal hyperparameters obtained by OptBA.}
  \begin{tabular}{@{}lcccc@{}}
    \toprule
    Settings & Precision & Recall & F1-score & Accuracy\\
    \midrule
    Default & 0.9837 & 0.9816 & 0.9816 & 98.19\% \\
    \midrule
    Optimal & 0.9887 & 0.9887 & 0.9891 & 99.63\% \\    
    \bottomrule
  \end{tabular}
  \label{tab:classification report}
\end{table*}

\begin{table*}[!t]
  \centering
   \caption{Comparison of OptBA against Optuna \cite{akiba2019optuna}.}
   \resizebox{\textwidth}{!}{%
    \begin{tabular}{@{}lcccccc@{}}
    \toprule
    Framework & Initial accuracy\% & Best accuracy\% & Best no. epochs & Best no. units & Parallel pruning & Profound search\\
    \midrule
    Optuna & 95.95 & 99.26 & 47 & 94 & No & No\\
    OptBA (ours) & 99.63 & 99.63 & 49 & 108 & Yes & Yes\\
    \bottomrule
  \end{tabular}
  \label{tab:optuna_optba}
  }
\end{table*}

In this section, we explore the augmented version of the ailment classification dataset \cite{Mooney}, and we compare the proposed algorithm with SOTA. First, data preprocessing techniques were applied. Then, textual data were transformed into a numerical format using the word embedding technique, where each word is represented by a vector of size 32. Finally, we applied 10-fold cross-validation along with LSTM to predict the patient's ailment. All experiments were run using Quadro RTX 6000 GPU with 24GB.

\subsection{Model Configuration}
The LSTM architecture includes one LSTM layer with 64 units followed by a dropout rate of 0.2. The default training details involve 20 epochs, a batch size of 10, and a learning rate of $1e-3$. Additionally, the hyperparameters of OptBA consist of a population size ($n$) of 10, a good population size ($m$) of 7, an elite population size ($e$) of 3, with 4 elite bees recruited ($nep$), and 1 each for the number of good bees recruited ($nsp$) and neighbourhood size ($ngh$).


To get the highest possible accuracy, we implemented OptBA to acquire the ideal hyperparameters for the LSTM model. Table~\ref{tab:classification report} indicates that the performance increased when the output dimensionality (the number of units) is adjusted to 108 along with running 49 epochs, which in return, increased the accuracy score by approximately 1.4\% compared with the baseline model. It is crucial to contextualize this improvement within the domain and dataset characteristics. As mentioned, our study is conducted on a novel dataset for which, to the best of our knowledge, no prior work exists. Therefore, even a marginal enhancement in accuracy can have significant implications, particularly in domains where precise classification of text data is critical, such as medical diagnosis. The performance of LSTM model was evaluated in Table~\ref{tab:classification report} based on well-known metrics for multi-class classification.


In order to extend the applicability of OptBA, we investigated its impact on Convolutional Neural Network (CNN). Specifically, we examined two key hyperparameters: the number of epochs and the number of kernels (also known as filters) in the convolutional layer. The results show a significant improvement where the accuracy increased by 3.3\% from 96.32\% with 20 epochs and 8 kernels to 99.63\% with 41 epochs and 51 kernels.


\subsection{Comparison with Optuna}
The architecture and the optimization techniques implemented by Optuna \cite{akiba2019optuna} can generate one optimal solution per trial. In contrast, OptBA employs an optimization method that is a population-based search, allowing the pruning of suboptimal solutions to occur concurrently for accelerated convergence. However, a direct comparison of total execution time is not feasible due to distinct numbers of solutions per trial and variations in search criteria between the two frameworks. For instance, OptBA swiftly identified the best solution in its initial trial and terminated early, whereas Optuna continued for 100 iterations without achieving the optimal outcome, as detailed in Table~\ref{tab:optuna_optba}. Moreover, the parameters of OptBA offer enhanced flexibility and depth in the quest for optimal solutions, without compromising the speed of evaluating an individual trial, but may require a larger number of trials. For example, setting $ngh=1$ entails a search extending one step forward and backwards from the current optimal solution. The lower the $ngh$ value, the more profound the search space. Consequently, OptBA guarantees to find the global optimal solution, distinguishing it from Optuna, which can get stuck in local optima as depicted in Table~\ref{tab:optuna_optba}.

\subsection{Limitations}
While the proposed OptBA algorithm shows promising results in optimizing hyperparameters for LSTM and CNN in medical text classification, its applicability to other domains or other deep learning models remains unexplored. Additionally, generalizing the success of OptBA to diverse datasets and larger hyperparameter search space requires further investigation.

\section{Conclusion}
One of the drawbacks of deep learning models is that they require much effort in tuning hyperparameters. Therefore, the proposed work introduced a novel mechanism in order to obtain the optimal hyperparameters required for building deep neural networks without getting stuck in local optima. This mechanism utilizes the Bees Algorithm\textemdash one of the recent swarm intelligence algorithms that is adapted to work on LSTM for the aim of classifying ailments based on medical text. Experiments indicated that the proposed framework could produce promising results and significantly improve the performance of deep neural networks. For future research, this work can be extended to explore other domains as well as datasets and other deep learning models.

\bibliography{main}

\end{document}